\pdfoutput=1
\UseRawInputEncoding
\documentclass[11pt]{article}

\usepackage[review]{acl}
\usepackage{times}
\usepackage{latexsym}
\usepackage{color}
\usepackage{bbding}
\usepackage{svg}
\usepackage{amsmath}
\usepackage{microtype}
\usepackage[misc]{ifsym}
\usepackage{bm}
\usepackage[utf8]{inputenc}
\usepackage{mdwlist}
\usepackage{float}
\usepackage{multirow}
\usepackage{CJK}
\usepackage{graphicx}  %插入图片的宏包
\usepackage{booktabs}

\title{DABERT: Dual Attention Enhanced BERT for Semantic Matching}

\author{Sirui Wang\textsuperscript{1,*} , Di Liang\textsuperscript{1,2,†} ,  Jian Song\textsuperscript{†} ,  Yuntao Li \textsuperscript{†}, Wei Wu\textsuperscript{†}\\
  Tsinghua University, Beijing, China\textsuperscript{*}  \\
  Meituan Inc.,Beijing, China\textsuperscript{†} \\
  \texttt{\{wangsirui,liangdi04,songjian20,liyuntao04,wuwei30\}@meituan.com} \\}

\begin{document}

\maketitle
\footnotetext[1]{These authors contributed equally to this work.} 
\footnotetext[2]{Corresponding author.}

\begin{abstract}
Transformer-based pre-trained language models such as BERT have achieved remarkable results in Semantic Sentence Matching. However, existing models still suffer from insufficient ability to capture subtle differences.
Minor noise like word addition, deletion, and modification of sentences may cause flipped predictions.
To alleviate this problem, we propose a novel \textbf{Dual Attention Enhanced BERT (DABERT)} to enhance the ability of BERT to capture fine-grained differences in sentence pairs. DABERT comprises (1) Dual Attention module, which measures soft word matches by introducing a new dual channel alignment mechanism to model affinity and difference attention. (2) Adaptive Fusion module, this module uses attention to learn the aggregation of difference and affinity features, and generates a vector describing the matching details of sentence pairs.
We conduct extensive experiments on well-studied semantic matching and robustness test datasets, and the experimental results show the effectiveness of our proposed method. 
\end{abstract}

\section{Introduction}

Semantic Sentence Matching (SSM) is a fundamental NLP task. The goal of SSM is to compare two sentences and identify their semantic relationship. In paraphrase identification, SSM is used to determine whether two sentences are paraphrase or not \cite{madnani-etal-2012-examining}.
In natural language inference task, SSM is utilized to judge whether a hypothesis sentence can be inferred from a premise sentence \cite{bowman2015large}. 
In the answer sentence selection task, SSM is employed to assess the relevance between query-answer pairs and rank all candidate answers.

\begin{figure}
\centering
\includegraphics[width=0.48\textwidth]{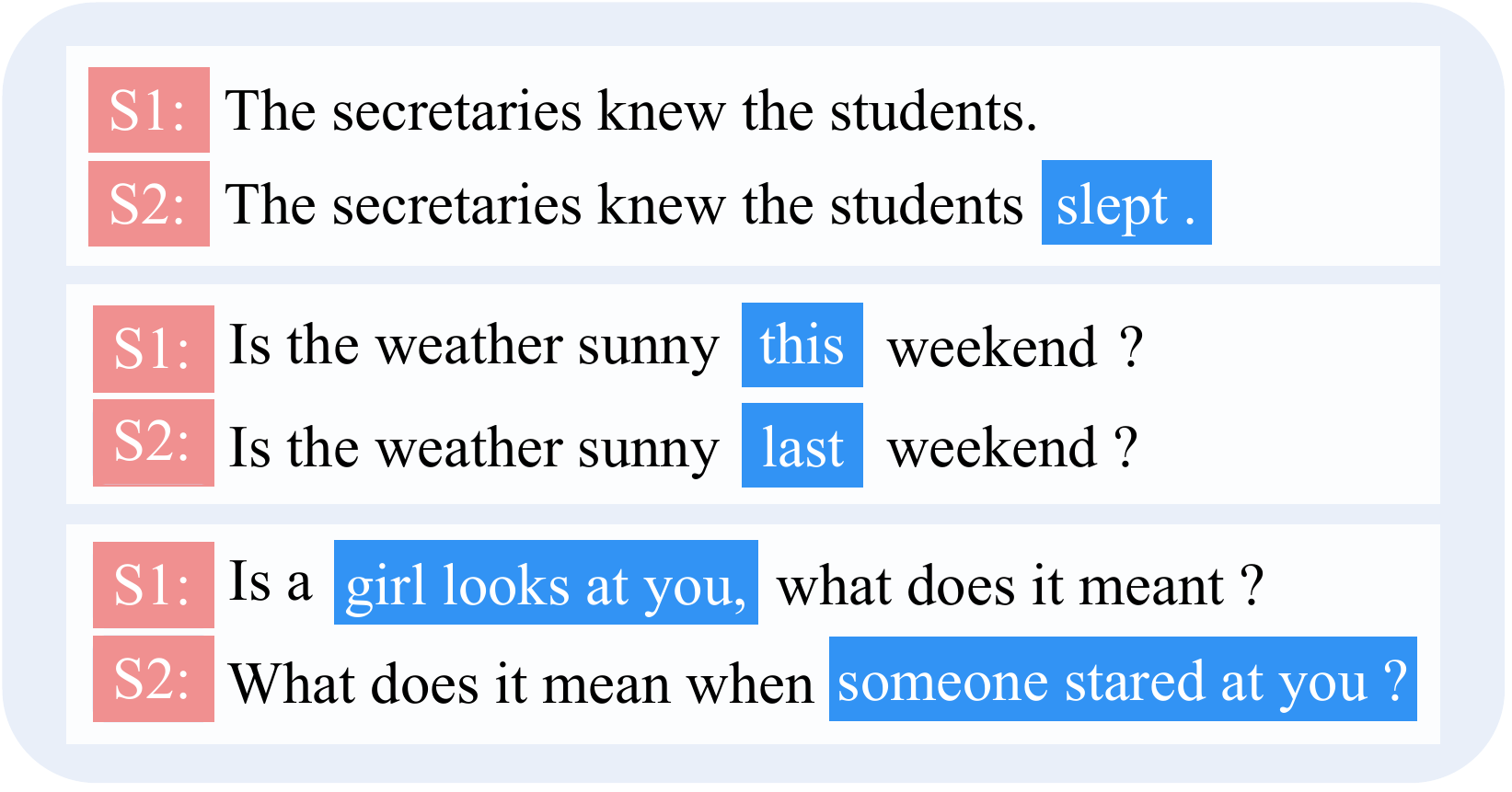}
\caption{\label{fig:example}Example sentences with similar text but different semantics. S1 and S2 are sentence pair. }
\end{figure}

Across the rich history of semantic sentence matching research, there have been two main streams of studies for solving this problem. One is to utilize a sentence encoder to convert sentences into low-dimensional vectors in the latent space, and apply a parameterized function to learn the matching scores between them \cite{reimers2019sentence}. 
Another paradigm adopts attention mechanism to calculate scores between tokens from two sentences, and then the matching scores are aggregated to make a sentence-level decision \cite{chen2016enhanced,tay2017compare}.
In recent years, pre-trained models, such as BERT~\cite{devlin2018bert}, RoBERTa~\cite{liu2019roberta}, have became much more popular and achieved outstanding performance in SSM. Recent work also attempts to enhance the performance of BERT by injecting knowledge into it, such as SemBERT \cite{zhang2020semantics}, UER-BERT \cite{xia2021using}, Syntax-BERT \cite{bai2021syntax} and so on.

Although previous studies have provided some insights, those models do not perform well in distinguishing sentence pairs with high literal similarities but different semantics. Figure \ref{fig:example} demonstrates several cases suffering from this problem.  Although the sentence pairs in this figure are semantically different, they are too similar in literal for those pre-trained language models to distinguish accurately. This could be caused by the 
self-attention architecture itself. 
Self-attention mechanism focuses on using the context of a word to understand the semantics of the word, while ignoring modeling the semantic difference between sentence pairs.
De-attention~\cite{NEURIPS2019_16fc18d7} and Sparsegen~\cite{martins2016softmax} have proved that equipping with attention mechanism with more flexible structure, models can generate more powerful representations. In this paper, we also focus on enhancing the attention mechanism in transformer-based pre-trained models to better integrate difference information between sentence pairs.
We hypothesize that paying more attention to the fine-grained semantic differences, explicitly modeling the difference and affinity vectors together will further improve the performance of pre-trained model. 
Therefore, two systemic questions arise naturally:

\textbf{Q1: How to equip vanilla attention mechanism with the ability on modeling semantics of fine-grained differences between a sentence pair?}
Vanilla attention, or named affinity attention, less focuses on the fine-grained difference between sentence pairs, which may lead to error predictions for SSM tasks.
An intuitive solution to this problem is to make subtraction between representation vectors to harvest their semantic differentiation. In this paper, we propose a dual attention module including a difference attention accompanied with the affinity attention. The difference attention uses subtraction-based cross-attention to aggregate word- and phrase- level interaction differences. Meanwhile, to fully utilize the difference information, we use dual-channel inject the difference information into the multi-head attention in the transformer to obtain semantic representations describing affinity and difference respectively.

\textbf{Q2: How to fuse two types of semantic representations into a unified representation?}
A hard fusion of two signals by extra structure may break the representing ability of the pre-trained model. How to inject those information softly to pre-trained model remains a hard issue. In this paper, we propose an Adaptive Fusion module, which uses an additional attention to learn the difference and affinity features to generate vectors describing sentence matching details. It first inter-aligns the two signals through distinct attentions to capture semantic interactions, and then uses gated fusion to adaptively fuse the difference features. Those generated vectors are further scaled with another fuse-gate module to reduce the damage of the pre-trained model caused by the injection of difference information. The output final vectors can better describe the matching details of sentence pairs. 

Our main contributions are three fold:

\begin{itemize}
\setlength{\itemsep}{1pt}
\setlength{\parsep}{0pt}
\setlength{\parskip}{0pt}
\item 
We point out that explicitly modeling fine-grained difference semantics between sentence pairs can effectively benefit sentence semantic matching tasks, and we propose a novel dual attention enhanced mechanism based on BERT.

\item 
Our proposed DABERT model uses a dual-channel attention to separately focus on the affinity and difference features in sentence pairs, and adopts a soft-integrated regulation mechanism to adaptively aggregate those two features. Thereby, the generated vectors can better describe the matching details of sentence pairs.
\item
To verify the effectiveness of DABERT, we conduct experiments on 10 semantic matching datasets and several data-noised dataset to test model's robustness. The results show that DABERT achieves an absolute improvement for over 2\% compared with pure BERT and outperforms other BERT-based models with more advanced techniques and external data usage.
\end{itemize}

\section{Approach}
\begin{figure*}
\centering
\includegraphics[width=0.95\textwidth]{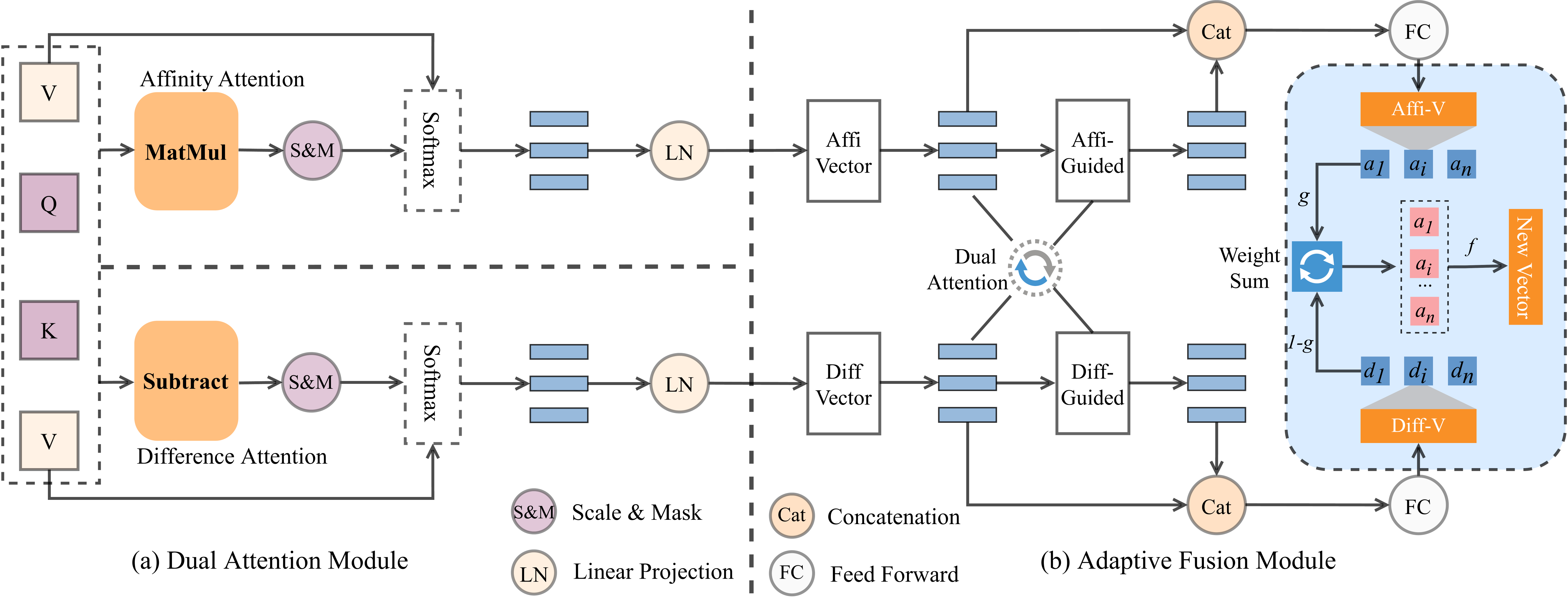}
\caption{\label{fig:overview_new} The overall architecture of Dual Attention Enhanced BERT (DABERT). The left side is the Dual attention module, and the right side is the Adaptive Fusion module.
}
\end{figure*}

Our proposed DABERT is a modification of the original transformer structure, whose structure is shown in Figure \ref{fig:overview_new}. Two submodules are included in this new structure. (1) Dual Attention Module, which uses a dual channel mechanism in multi-head attention to match words between two sentences. Each channel uses a different attention head to calculate affinity and difference scores separately, and obtains two representations to measure affinity and difference information respectively. (2) Adaptive Fusion Module, which is used to fuse the representation obtained by dual attention. It first uses guide-attention to align the two signals. And then, multiple gate modules are used to fuse the two signals. Finally, a vector is output including more fine-grained matching details.
In the following sections, we explain each component in detail.

\subsection{Dual Attention Module}

In this module, we use two distinct attention functions, namely affinity attention and difference attention, to compare the affinities and differences of vectors between two sentences. 
The input of the dual attention module is a triple of $K, Q, V \in R^{d_{seq} \times d_v}$, where $d_{v}$ is the latent dimension and $d_{seq}$ is the utterance length. Dual attention module calculate the latent relationship between $K, Q$ and $V$ via two separate attention mechanism to measure their affinity and difference.
As a result, two set of attention representations are generated by the dual attention module, which will be fused by the following adaptive fusion module.

\subsubsection{Affinity Attention}

The affinity attention module is the part of the dual attention module, which is the standard dot-product attention that operates following Transformer's default operation.
The input of affinity attention module  consists of queries and keys of dimension $d_k$, and values of dimension $d_v$. We compute the dot products of the query with all keys, divide each by $\sqrt{d_k}$, and apply a softmax function to obtain the weights on the values. For the sake of simplicity, the formulations of BERT not be repeated here, please refer to \cite{devlin2018bert} for more details. We denote the output affinity vector as:

\begin{small}
\begin{equation}
    \begin{aligned}
    \mathbf{A} = softmax(\frac{\mathbf{Q K}^T}{\sqrt{d_k}}) * \mathbf{V},
    \end{aligned}
\end{equation}
\end{small}
where $\mathbf{A}=\{\bm{a}_1,...,\bm{a}_l\}\in R^{{d_l} \times d_{v}}$ denotes the vector describing affinity expressions generated by the Transformer original attention module.

\subsubsection{Difference Attention}
The second part of dual attention module is a difference attention module that capture and aggregate the difference information between sentence pairs. The difference attention module adopts a subtraction-based cross-attention mechanism, which allows model to pay attention to dissimilar parts between sentence pairs by element-wise subtraction as:

\begin{small}
\begin{gather}
\mathbf{D} = softmax(\frac{\beta}{\sqrt{d_k}}) * \mathbf{V}, \\
\bm{\beta} = \left\|\mathbf{Q}-\mathbf{K}\right\|+\mathbf{M}, \\
\left\|\mathbf{Q} - \mathbf{K}\right\|_{ij} = \sum_{k=0}^{d_k} \mathbf{Q}_{ik} - \mathbf{K}_{jk}, 
\end{gather}
\end{small}
where $\left\|\mathbf{Q}-\mathbf{K}\right\| \in R^{d_{l} \times d_{l}}$
and $d_{l}$ is the input sequence length. We use $\mathbf{D}=\{\bm{d}_1,...,\bm{d}_l\}\in R^{{d_l} \times d_{v}}$ to denote the representation generated by the difference attention. The $\mathbf{M} \in R^{d_{l} \times d_{l}} $ is a masking operation.
Both the affinity attention and the difference attention are utilized to fit the semantic relationship of sentence pairs, and obtain the representations with the same dimension from the perspective of affinity and difference respectively. This dual channel mechanism can obtain more detailed representations describing sentence matching.

\subsection{Adaptive Fusion Module}

After obtaining the affinity signals $\mathbf{A}$ and the difference signals $\mathbf{D}$, we introduce a novel adaptive fusion module to fuse these two signals instead of direct fusion (i.e., average embedding vector), since direct fusion may compromise the original representing ability of the pre-trained model. The fusion process includes three steps.
First, it flexibly interacts and aligns these two signals via affinity-guided attention and difference-guided attention. Second, multiple gate modules are adopted to selectively extract interaction semantic information. Finally, to alleviate the damage of the pre-trained model by the difference signal, 
we utilize filter gates to adaptively filter out noisy information and finally generate vectors that better describe the details of sentence matching.

Firstly, we update the difference vectors through affinity-guided attention. We use $\bm{a}_i$ and $\bm{d}_i$ to denote the $i$-th dimension of $\mathbf{A}$ and $\mathbf{D}$ respectively. 
We provide each affinity vector $\bm{a}_i$ to interact with the difference signals matrix $\mathbf{D}$ and obtain the new difference feature $\bm{d}^*_i$. Then, based on $\bm{d}^*_i$, we can in turn acquire the new Affinity feature $\bm{a}^*_i$ through difference-guided attention. The calculation process is as follows:
\begin{normalsize}
\begin{equation}
    \begin{aligned}
        \delta_i &= \tanh(\mathbf{W}_{D} \mathbf{D} \oplus (\mathbf{W}_{a_i} a_i + b_{a_i})), \\
        \overline{d}_i &= \mathbf{D} * softmax(\mathbf{W}_{d_i} \delta_i + b_{d_i}), \\
\gamma_i &= \tanh(\mathbf{W}_{A} \mathbf{A} \oplus (\mathbf{W}_{\overline{d}_i} \overline{d}_i + b_{\overline{d}_i})), \\
\overline{a}_i &= \mathbf{A} * softmax(\mathbf{W}_{\overline{a}_i} \gamma_i + b_{\overline{a}*_i}),\\
d^*_i &= \tanh(\mathbf{W}_{d^*_i}( [d_i;\overline{d}_i] )+b_{d^*_i})),\\
a^*_i &= \tanh(\mathbf{W}_{a^*_i}([a_i;\overline{a}_i])+b_{a^*_i})),
    \end{aligned}
\end{equation}
\end{normalsize}
where ${\mathbf{W}_{D}, \mathbf{W}_{A}, \mathbf{W}_{a_i}, \mathbf{W}_{\overline{d}_i}}\in {R}^{d_{l}*{d_v}}$; $\mathbf{W}_{d_i}$, $\mathbf{W}_{\overline{a}_i}\in R^{1*2d_{l}}$; $ b_{d^*_i}$, $b_{a_i}$, $b_{d_i}$, $b_{a^*_i}$ are weights and bias of our model, and $\oplus$ denotes the concatenation of signal matrix and feature vector. 
Secondly, to adaptively capture and fuse useful information from Affinity and difference features, we introduce our gate fusion modules:
\begin{normalsize}
\begin{equation}
    \begin{aligned}
       \hat{d_i} &= \tanh(\mathbf{W}_{\hat{d_i}} d^*_i + b_{\hat{d_i}}), \\
\hat{a_i} &= \tanh(\mathbf{W}_{\hat{a_i}} a^*_i + b_{\hat{a_i}}), \\
g_i &= \sigma(\mathbf{W}_{g_i}(\hat{d_i} \oplus \hat{a_i})), \\
v_i &= g_i \hat{a_i} + (1 - g_i) \hat{d_i},
    \end{aligned}
\end{equation}
\end{normalsize}
where $\mathbf{W}_{\hat{d_i}}$, $\mathbf{W}_{\hat{a_i}}\in R^{d_{h}*{d_v}}$; $\mathbf{W}_{g_i}\in R^{1*2d_{h}}$; $b_{\hat{d_i}}$, $b_{\hat{a_i}}$ are parameters and $d_{h}$ is the size of hidden layer. $\sigma$ is the sigmoid activation function and $g_i$ is the gate that determines the transmission of these two distinct representations. By the way, we get the fusion feature $\bm{v}_i$ .

Eventually, considering the potential noise problem, we propose a filtering gate to selectively leverage the fusion feature. When $\bm{v}_i$ tends to be beneficial, the filtration gate will incorporate the fusion features and the original features. Otherwise, the fusion information will be filtered out:
\begin{normalsize}
\begin{equation}
    \begin{aligned}
        f_i &= \sigma(\mathbf{W}_{f_i,a_i} (a_i \oplus (\mathbf{W}_{v_i} v_i + b_{v_i} ))), \\
        l_i &= f_i * \tanh(\mathbf{W}_{l_i} v_i + b_{l_i}),
    \end{aligned}
\end{equation}
\end{normalsize}
where $\mathbf{W}_{f_i,a_i}\in R^{1*2{d_v}}$; $\mathbf{W}_{v_i}$, $\mathbf{W}_{l_i}\in R^{{d_v}*d_{h}}$; $b_{v_i}$, $b_{l_i}$ are trainable parameters and $\bm{l}_i$ is the final fused semantic feature and it will be propagated to the next computation flow.

\begin{table*}
\centering
\renewcommand\arraystretch{0.8}
\scalebox{0.78}{
\setlength{\tabcolsep}{2.6mm}{
\begin{tabular}{lcccccccc}
\toprule
\midrule
\multirow{2}*{Model} &\multirow{2}*{Pre-train} & \multicolumn{3}{c}{Sentence Similarity} &\multicolumn{3}{c}{Sentence Inference} &\multirow{2}*{Avg}\\  \cmidrule(r){3-8} 
         ~& &\text{MRPC} & \text{QQP} & \text{SST-B} &\text{MNLI-m/mm}  & \text{QNLI} & \text{RTE}  \\

\midrule
\text{BiMPM$\dagger$\cite{wang2017bilateral}} & \XSolidBrush & 79.6 & 85.0 & - & 72.3/72.1 & 81.4 & 56.4  & - \\
\text{CAFE$\dagger$\cite{tay2017compare}}& \XSolidBrush  & 82.4 & 88.0 & - & 78.7/77.9 & 81.5 & 56.8  & - \\
\text{ESIM$\dagger$\cite{chen2016enhanced}}& \XSolidBrush & 80.3 & 88.2 & - & - & 80.5 & -  & - \\
\text{Transformer$\dagger$\cite{vaswani2017attention}}& \XSolidBrush  & 81.7 & 84.4 & 73.6 & 72.3/71.4 & 80.3 & 58.0  & 74.53 \\
\midrule

\text{BiLSTM+ELMo+Attn$\dagger$\cite{devlin2018bert}} &\Checkmark  & 84.6 & 86.7 & 73.3 & 76.4/76.1 & 79.8 & 56.8  & 76.24 \\
\text{OpenAI GPT$\dagger$} &\Checkmark  & 82.3 & 70.2 & 80.0 & 82.1/81.4 & 87.4 & 56.0  & 77.06 \\
\text{UERBERT$\ddagger$\cite{xia2021using}} &\Checkmark  & 88.3 & 90.5 & 85.1 & 84.2/83.5 & 90.6 & 67.1 & 84.19 \\
\text{SemBERT$\dagger$\cite{zhang2020semantics}} &\Checkmark   & 88.2 & 90.2 & 87.3 & 84.4/84.0 & 90.9 & 69.3  & 84.90 \\
\midrule
\text{BERT-base$\ddagger$\cite{devlin2018bert}}&\Checkmark  & 87.2 & 89.0 & 85.8 & 84.3/83.7 & 90.4 & 66.4  & 83.83 \\
\text{SyntaxBERT-base$\dagger$\cite{bai2021syntax}}&\Checkmark  & \textbf{89.2} & 89.6 & 88.1 & 84.9/84.6 & 91.1 & 68.9  & 85.20 \\
\textbf{DABERT-base}$\ddagger$&\Checkmark  & 89.1 & \textbf{91.3} & \textbf{88.2}  & \textbf{84.9}/\textbf{84.7} & \textbf{91.4} & \textbf{69.5} & \textbf{85.58} \\
\midrule
\text{BERT-large$\ddagger$\cite{devlin2018bert}}&\Checkmark  & 89.3 & 89.3 & 86.5 & 86.8/85.9 & 92.7 & 70.1  & 85.80 \\
\text{SyntaxBERT-large$\dagger$\cite{bai2021syntax}}&\Checkmark  & \textbf{92.0} & 89.5 & 88.5 & 86.7/86.6 & 92.8 & 74.7  & 87.26 \\
\textbf{DABERT-large}$\ddagger$ &\Checkmark & 91.4 & \textbf{91.9} & \textbf{89.5}& \textbf{87.1}/\textbf{86.9} & \textbf{94.8} & \textbf{75.3}  & \textbf{88.12} \\
\midrule
\bottomrule
\end{tabular}}}
\caption{\label{citation-guide-gule}
The performance comparison of DABERT with other methods. We report Accuracy $\times$ 100 on 6 GLUE datasets. Methods with $\dagger$ indicate the results from their papers, while methods with $\ddagger$ indicate our implementation.
}
\vspace{-0.2cm}
\end{table*}

\begin{table}[th]
\centering
\renewcommand\arraystretch{0.9}
\scalebox{0.75}{
\setlength{\tabcolsep}{1.3mm}{
\begin{tabular}{lcccccc}
\toprule
\text{Model} & \text{SNLI} & \text{Sci} & \text{SICK} & \text{Twi} \\
\midrule
\text{ESIM$\dagger$\cite{chen2016enhanced}$\quad$} & 88.0 & 70.6 & - & - \\
\text{CAFE$\dagger$\cite{tay2017compare}$\quad$}  & 88.5 & 83.3 & 72.3 & - \\
\text{CSRAN$\dagger$\cite{tay2018co}$\quad$} & 88.7 & 86.7 & - & 84.0 \\
\midrule
\text{BERT-base$\ddagger$\cite{devlin2018bert}} & 90.7 & 91.8 & 87.2 & 84.8 \\
\text{UERBERT$\ddagger$\cite{xia2021using}} & 90.8 & 92.2 & 87.8 & 86.2 \\
\text{SemBERT$\dagger$\cite{zhang2020semantics}} & 90.9 & 92.5 & 87.9 & 86.8 \\
\text{SyntaxBERT-base$\dagger$\cite{bai2021syntax}} & 91.0 & 92.7 & \textbf{88.7} & 87.3  \\
\textbf{DABERT-base$\ddagger$} & \textbf{91.3} & \textbf{93.6} & 88.6 & \textbf{87.5} \\
\midrule
\text{BERT-large$\ddagger$\cite{devlin2018bert}} & 91.0 & 94.4 & 91.1 & 91.5 \\
\text{SyntaxBERT-large$\dagger$\cite{bai2021syntax}} & 91.3 & 94.7 & 91.4 & 92.1 \\
\textbf{DABERT-large$\ddagger$} & \textbf{91.5} & \textbf{95.3} & \textbf{92.5} & \textbf{92.3} \\
\bottomrule
\end{tabular}}}
\caption{\label{citation-guide-outsideGlue}
The performance comparison of DABERT with other methods on 4 popular datasets, including SNLI, Scitail(Sci), SICK and TwitterURL(Twi).}
\vspace{-0.5cm}
\end{table}

\section{Experimental Settings}
\subsection{Datasets}

\textbf{Semantic Matching.} 
We conduct experiments on 10 sentence matching datasets to evaluate the effectiveness of our method. The GLUE~\cite{wang2018glue} benchmark is a widely-used dataset in thie field, which includes tasks such as sentence pair classification, similarity and paraphrase detection, and natural language inference\footnote{https://huggingface.co/datasets/glue}. We conduct experiments on 6 sentence pair datasets (MRPC, QQP, STS-B, MNLI, RTE, and QNLI) from GLUE. We also conduct experiments on 4 other popular datasets (SNLI \cite{bowman2015large}, SICK \cite{marelli2014sick}, TwitterURL \cite{lan2017continuously} and Scitail \cite{khot2018scitail}). The statistics of all 10 datasets are shown in Table \ref{citation-guide-new}.

\noindent \textbf{Robustness Test.}  
TextFlint \cite{gui2021textflint} is a robustness evaluation platform for natural language processing models\footnote{https://www.textflint.io}. It includes more than 80 patterns to deform data, including inserting punctuation marks, changing numbers in text, replacing synonyms, modifying adverbs, deleting words, etc. It can effectively evaluate the robustness and generalization of models. In this paper, we leverage TextFlint to perform transformations on multiple datasets (Quora, SNLI, MNLI-m/mm), including task-specific transformations (SwapAnt, NumWord, AddSent) and general transformations (InsertAdv, Appendlrr, AddPunc, BackTrans, TwitterType, SwapNamedEnt, SwapSyn-WordNet). We conduct experiments on datasets with those types of transformations to verify the robustness of our model.

\subsection{Baselines}
%\noindent \textbf{Baselines} 
To evaluate the effectiveness of our proposed DABERT in SSM, we mainly introduce BERT \cite{devlin2018bert}, SemBERT \cite{zhang2020semantics}, SyntaxBERT , UERBERT \cite{xia2021using} and multiple other PLMs \cite{devlin2018bert} for comparison. In addition, we also select several competitive models without pre-training as baselines, such as ESIM \cite{chen2016enhanced}, Transformer \cite{vaswani2017attention} , etc \cite{hochreiter1997long,wang2017bilateral,tay2017compare}. In robustness experiments, we compare the performance of multiple pre-trained models \cite{sanh2019distilbert,chen2016enhanced,devlin2018bert,lan2019albert} and SemBERT,UERBERT and Syntax-BERT on the robustness test datasets. For simplicity, the compared models are not described in detail here.

\subsection{Implementation Details}
DABERT is based on BERT-base and BERT-large. For distinct targets, our hyper-parameters are different.  We use AdamW in the BERT and set the learning rate in \{$1e^-5$, $2e^-5$, $3e^-5$, $8e^-6$\}. As for the learning rate decay, we use a warmup of 0.1 and  L2 weight decay of 0.01. Furthermore, we set the epoch to 5 and the batch size is selected in \{16, 32, 64\}. We also set dropout at 0.1-0.3. To prevent gradient explosion, we set gradient clipping in \{7.5, 10.0, 15.0\}. All the experiments are conducted by Tesla V100 and PyTorch platform. In addition, to ensure that the experimental results are statistically significant, we conduct each experiment five times and report the average results.

\begin{table*}[t]
	\centering
	\renewcommand\arraystretch{0.9}
    \setlength{\tabcolsep}{1mm}
	{   \scalebox{0.78}{
	\setlength{\tabcolsep}{2.6mm}
		\begin{tabular}{lcccccccccc}
		\toprule
		\midrule
		\multirow{2}*{Model} &\multicolumn{5}{c}{Quora} &\multicolumn{5}{c}{SNLI} \\  \cmidrule(r){2-6} \cmidrule(r){7-11} 
         ~ &SA &NW &IA &Al &BT \quad\quad &AS &SA &TT &SN &SW \\
        \midrule
        ESIM$\dagger$\cite{chen2016enhanced}\quad&-& -&-&-&- \quad\quad&64.00& 84.22&78.32&53.76&65.38 \\
        %DistilBERT$\dagger$\cite{sanh2019distilbert}\quad\quad
        %& 42.24& 56.85&83.10&84.09&83.20 \quad\quad
        %&-& -&-&-&- \\
        BERT$\ddagger$\cite{devlin2018bert}\quad\quad
        &48.58&56.96&86.32&\textbf{85.48}&83.42 \quad\quad
        &79.66&94.84&83.56&50.45&76.42 \\
        ALBERT$\ddagger$\cite{lan2019albert}\quad\quad
        &51.08&55.24&81.87&78.94&82.37 \quad\quad
        &45.17&96.37&81.62&57.66&74.93 \\
        UERBERT$\ddagger$\cite{xia2021using}\quad\quad
        &48.57&54.86&84.72&80.88&82.71 \quad\quad
        &73.24&94.78&85.36&57.54&80.81 \\
        SemBERT$\ddagger$\cite{zhang2020semantics}\quad\quad
        &50.92&53.15&85.19&82.04&82.40 \quad\quad
        &76.81&95.31&84.60&56.28&77.86 \\
        SyntaxBERT$\ddagger$\cite{bai2021syntax}\quad\quad
        &49.30&56.37&86.43&84.62&84.19 \quad\quad
        &78.63&95.02&\textbf{86.91}&58.26&76.90 \\
        \midrule
        \textbf{DABERT}$\ddagger$ & \textbf{60.43}& \textbf{62.76}&\textbf{87.50}&85.48&\textbf{87.49} \quad\quad
        &\textbf{81.06}&\textbf{96.85}&85.14&\textbf{60.58}&\textbf{80.92} \\
		\end{tabular}}	
		\renewcommand\arraystretch{0.9}
		\scalebox{0.78}{
		\setlength{\tabcolsep}{3.1mm}
	    \begin{tabular}{lcccccc}
		\toprule
		\multirow{2}*{Method} &\multicolumn{6}{c}{MNLI-m/mm} \\  \cmidrule(r){2-7}
         ~ &AS &SA &AP &TT &SN &SW \\
        \midrule
        BERT$\ddagger$\cite{devlin2018bert}
        &55.32/55.25&52.76/55.69&82.30/82.31&77.08/77.22&51.97/51.84&76.41/77.05 \\
        ALBERT$\ddagger$\cite{lan2019albert}
        &53.09/53.58&50.25/50.20&\textbf{83.98/83.68}&\textbf{77.98}/78.03&56.43/50.03&76.63/77.43 \\
        UERBERT$\ddagger$\cite{xia2021using}
        &54.99/54.84&52.29/53.80&79.80/79.18&75.46/74.93&55.21/55.96&\textbf{82.23}/82.74 \\
        SemBERT$\ddagger$\cite{zhang2020semantics}
        &55.38/55.12&54.07/54.62&78.70/78.16&73.90/73.47&53.43/53.76&78.09/78.93 \\
        SyntaxBERT$\ddagger$\cite{bai2021syntax}
        &54.92/54.63&53.54/54.73&77.01/76.71&70.38/70.13&57.11/51.95&78.57/79.31 \\
        \midrule
        \textbf{DABERT}$\ddagger$ & \textbf{60.14/59.25}& \textbf{60.89/61.37}&83.23/83.19&77.94/\textbf{78.10}&\textbf{60.12/59.83}&82.15/\textbf{82.97} \\
        \bottomrule
        \midrule
		\end{tabular}}
    }\caption{\label{citation-guide-robust}The robustness experiment results of DABERT and other models. The data transformation methods we utilized mainly include SwapAnt (SA), NumWord (NW), AddSent (AS), InsertAdv (IA), Appendlrr (Al), AddPunc (AP), BackTrans (BT), TwitterType (TT), SwapNamedEnt (SN), SwapSyn-WordNet (SW).}
\end{table*}

\section{Results and Analysis}
\subsection{Model Performance}
In our experiments, we implement DABERT in the initial transformer layer of BERT.

First, we fine-tune our model on 6 GLUE datasets. Table \ref{citation-guide-gule} shows the performance of DABERT and other competitive models. It can be seen that using only non-pretrained models performs obviously worse than PLMs due to their strong context awareness and data fitting capabilities.
When the backbone model is BERT-base or BERT-large, the average accuracy of DABERT respectively improves by 1.7\% and 2.3\% than vanilla BERT. Such great improvement demonstrates the benefit of fusion difference attention for mining semantics and proves that our framework can help BERT perform much better in SSM.

Moreover, compared with some previous works such as SemBERT, UERBERT and SyntaxBERT, DABERT achieves the best performance without injecting external knowledge. 
Specifically, our model outperforms SyntaxBERT, the best performing model in previous work leveraging external knowledge, with an average relative improvement of 0.86\% based on BERT-large. On the QQP dataset, the accuracy of DABERT is significantly improved by 2.4\% over SyntaxBERT. There are two main reasons for such results. 
On the one hand, we use dual-channel attention to enhance the ability of DABERT to capture difference features. This enables DABERT to obtain more fine-grained interaction matching features.
On the other hand, for the potential noise problem introduced by external structures, our adaptive fusion module can selectively filter out inappropriate information to suppress the propagation of noise, and previous work does not seem to pay enough attention to this problem. 
However, we still notice that SyntaxBERT achieves slightly better accuracy on a few datasets. We argue that this is a result of the intrinsic correlation of syntactic and dependent knowledge.

Second, to verify the general performance of our method, we also conduct experiments on other popular datasets. The results are shown in Table \ref{citation-guide-outsideGlue}. DABERT still outperforms vanilla BERT and other models on almost all datasets. 
 It is worth noting that DABERT performs worse than SyntaxBERT on SICK. 
 This may be because the data volume of SICK is relatively small, and SyntaxBERT uses syntactic prior knowledge, which makes SyntaxBERT more advantageous on small datasets. 
 but DABERT still shows a very competitive performance on SICK, which also shows from the side that our method can enhance the difference capture ability of BERT and make up for the lack of generalization ability with fewer parameters.
 
Overall, our method has competitive performance in judging semantic similarity compared to previous work. Extensive performance improvements also validate our point,
soft ensemble difference information based on BERT's powerful contextual representation capability is useful for sentence matching tasks.

\begin{table*}[t]
\centering
\renewcommand\arraystretch{0.95}
\scalebox{0.88}{
\setlength{\tabcolsep}{3mm}{
\begin{tabular}{lcccc}
\toprule
\midrule
\text{Case} & \text{ESIM} & \text{BERT} & \text{SyntaxBERT}& \text{DABERT}  \\
\midrule
\text{S1:}How done \textcolor{red}{you solve} this aptitude question? &
\multirow{2}{*}{\text{label:1}}& \multirow{2}{*}{\text{label:0}} & \multirow{2}{*}{\text{label:0}}& similarity:10.87\% \\
\text{S2:}How does \textcolor{blue}{I solve} aptitude questions \textcolor{blue}{on cube}? &  &   & &label:0 \\
\midrule
\text{S1:}How can I tell if \textcolor{red}{ this girl loves} me? &
\multirow{2}{*}{\text{label:1}}& \multirow{2}{*}{\text{label:1}} & \multirow{2}{*}{\text{label:1}}& similarity:12.06\% \\
\text{S2:}How can I tell if \textcolor{blue}{ this boy loves} me? &   &  &  & label:0\\
\midrule
\text{S1:}How many \textcolor{red}{12 digits number} have the sum of 4? &
\multirow{2}{*}{\text{label:1}}& \multirow{2}{*}{\text{label:1}} & \multirow{2}{*}{\text{label:1}}& similarity:18.63\% \\
\text{S2:}How many \textcolor{blue}{42 digits number} have the sum of 4? &   & 
& &label:0 \\
\midrule
\bottomrule
\end{tabular}}}
\caption{\label{citation-guide-casestudy2} The example sentence pairs of our cases. \textcolor{red}{Red} and \textcolor{blue}{Blue} are difference phrases in sentence pair. 
}
\vspace{-0.2cm}
\end{table*}

\subsection{Robustness Test Performance}

In order to examine the performance of DABERT and competitive models in their ability to capture subtle differences in sentence pairs. We perform robustness tests on three extensively studied datasets.

Table \ref{citation-guide-robust} lists the accuracy of DABERT and six baseline models on the three datasets. We can observe that SwapAnt leads to a drop in maximum performance, and our model outperforms the best model SemBert nearly 10\% on SwapAnt(QQP), which indicates that DABERT can better handle semantic contradictions caused by antonyms than baseline models.
And the model performance drops to 56.96\% on NumWord transformation, while DABERT outperforms BERT by nearly 6\% because it requires the model to capture subtle numerical differences for correct linguistic inference.
In SwapSyn transformation, UERBERT significantly outperforms other baseline models because it explicitly uses the synonym similarity matrix to calibrate the attention distribution, while our model can still achieve comparable performance to UERBERT without adding external knowledge.
On TwitterType and AddPunc, the performance of SyntaxBERT by injecting syntax trees degrades significantly, probably because converting text to twitter type or adding punctuation breaks the normal syntactic structure of sentences. And DABERT still achieves the competitive performance in these two transformations.
In other scenarios, DABERT also achieve better performance due to capturing subtle differences in sentence pairs. 
Meanwhile, ESIM has the worst performance, the results reflect that the pre-training mechanism benefits from rich external resources and provides better generalization ability than de novo trained models.
And the improved pre-trained model SyntaxBERT performs better than the original BERT model, which reflects that sufficient pre-trained corpus and suitable external knowledge fusion strategies can help improve the generalization performance of the model.

\begin{table}
\centering
\renewcommand\arraystretch{0.95}
\scalebox{0.78}{
\setlength{\tabcolsep}{3.5mm}{
\begin{tabular}{lcc|cc}
\toprule
%\hline
\multirow{2}*{Model} &\multicolumn{2}{c}{Quora} &\multicolumn{2}{c}{QNLI} \\ 
  \cmidrule(r){2-5}
  ~ & \text{Dev} & \text{Test} & \text{Dev} & \text{Test} \\
\midrule
%\hline
\textbf{DABERT} & \textbf{92.1} & \textbf{91.3}  & \textbf{92.9} & \textbf{91.4}\\
\text{w/o \ \ Affi-attention} & 90.1 & 89.5  & 91.8 & 90.7\\
\text{w/o \ \ Diff-attention} & 90.6 & 89.8  & 92.0 & 90.8\\
\text{w/o \ \ Guide-attention} & 91.3 & 90.4  & 92.1 & 91.0\\
\text{w/o \ \ Gate fusion} & 91.7 & 90.6  & 92.5 & 91.1\\
\text{w/o \ \ Gate filter} & 91.8 & 90.9 & 92.6 & 91.2 \\
\text{w/o \ \ Regulation} & 89.9 & 89.4 & 91.5 & 90.7 \\
\bottomrule
%\hline
\end{tabular}}}
\caption{\label{citation-guide-ablation}
Results of component ablation experiment.
}
\vspace{-0.45cm}
\end{table}

\subsection{Ablation Study}

To evaluate the contribution of each component in our method, we conduct ablation experiments on the QQP and QNLI datasets based on BERT. The experimental results are shown in the table \ref{citation-guide-ablation}.

Above all, the dual attention module consists of two core components that use a two-channel mechanism to model affinity and difference attention.
First, after removing affinity attention, the performance of the model on the two datasets drops by 1.8\% and 0.7\%. Affinity attention can capture the dynamic alignment relationship between word pairs, which is crucial for SSM tasks. Next,after removing difference attention from the model, the performance on the two datasets dropped by 1.5\% and 0.6\%, respectively. The difference information can further describe the interaction between words, and can provide more fine-grained comparison information for the pre-trained model, so that the model can obtain a better representation. The above experiments show that the performance drops sharply after the submodule is removed, which demonstrates the effectiveness of the internal components of the dual attention module.

Next, in the adaptive fusion module, we also conducted several experiments to verify the effect of the fusion of affinity and difference vectors. On the QQP dataset, we first remove the guide attention module, and the performance drops to 90.4\%. Since guide attention can capture the interaction between two signals, this interaction information is crucial for fusing two different information.
Second, after removing the fusion gate, we only integrate two signals by simple averaging. The accuracy dropped to 91.4\%, indicating that dynamically merging the affinity and difference vectors according to different weights can improve the performance of the model.
Then, when the filter gate is removed, the accuracy drops by 0.4\%, indicating that the ability of the model to suppress noise is weakened without the filter gate.
Finally, we also replaced the overall aggregation and Regulation module with simple average, and the performance dropped sharply to 89.4\%. While difference information is crucial for judging sentence-pair relations, hard-integrating the difference information into the PLMs will destroy its Pre-existing knowledge, and soft aggregation and governance can make better use of difference signals.

Overall, due to the effective combination of each component, DABERT can adaptively fuse difference features into pretrained models and leverage its powerful contextual representation to better inference about semantics.

\begin{table}
\centering
\renewcommand\arraystretch{0.83}
\setlength{\tabcolsep}{0.7mm}{
\scalebox{0.75}{
\setlength{\tabcolsep}{2mm}{
\begin{tabular}{lccccc}
\toprule
\text{Datasets} & \text{\#Train} & \text{\#Dev} & \text{\#Test} & \text{\#Class} \\
\midrule
\text{MRPC} & 3669 & 409 & 1380 & 2 \\
\text{QQP} & 363871 & 1501 & 390965 & 2 \\
\text{MNLI-m/mm} & 392703 & 9816/9833 & 9797/9848 & 3 \\
\text{QNLI} & 104744 & 40432 & 5464 & 2 \\
\text{RTE} & 2491 & 5462 & 3001 & 2 \\
\text{SST-B} & 5749 & 1500 & 1379 & 2 \\
\text{SNLI} & 549367 & 9842 & 9824 & 3 \\
\text{SICK} & 4439 & 495 & 4906 & 3 \\
\text{Scitail} & 23596 & 1304 & 2126 & 2 \\
\text{TwitterURL} & 42200 & 3000 & 9324 & 2 \\
\bottomrule
\end{tabular}}}}
\caption{\label{citation-guide-new}
The statistics of all 10 datasets.
}
\vspace{-0.5cm}
\end{table}

\subsection{Case Study}
To visualize how DABERT works, we use three cases from the table \ref{citation-guide-casestudy2} for qualitative analysis. In the first case, the non-pretrained language model ESIM is difficulty capturing the semantic conflicts caused by the difference words. Therefore, ESIM gives wrong prediction results in case 1.
BERT can identify the semantic difference in case 1 with the help of context representation %, so it makes the correct prediction in case 1
. But in case 3, BERT cannot capture the difference between the numbers "12" and "24" and give wrong prediction.
SyntaxBERT enhances text understanding by introducing syntactic trees. Since case 2 and case 3 have the same syntactic structure, SyntaxBERT also gives wrong predictions.
Our model made correct predictions in all of the above cases. Because DABERT explicitly focuses on different parts of sentence pairs through difference attention and adaptively aggregates affinity and difference information in the adaptive fusion module, it can identify semantic differences caused by subtle differences within sentence pairs.

\begin{figure}
\centering
\includegraphics[width=0.42\textwidth]{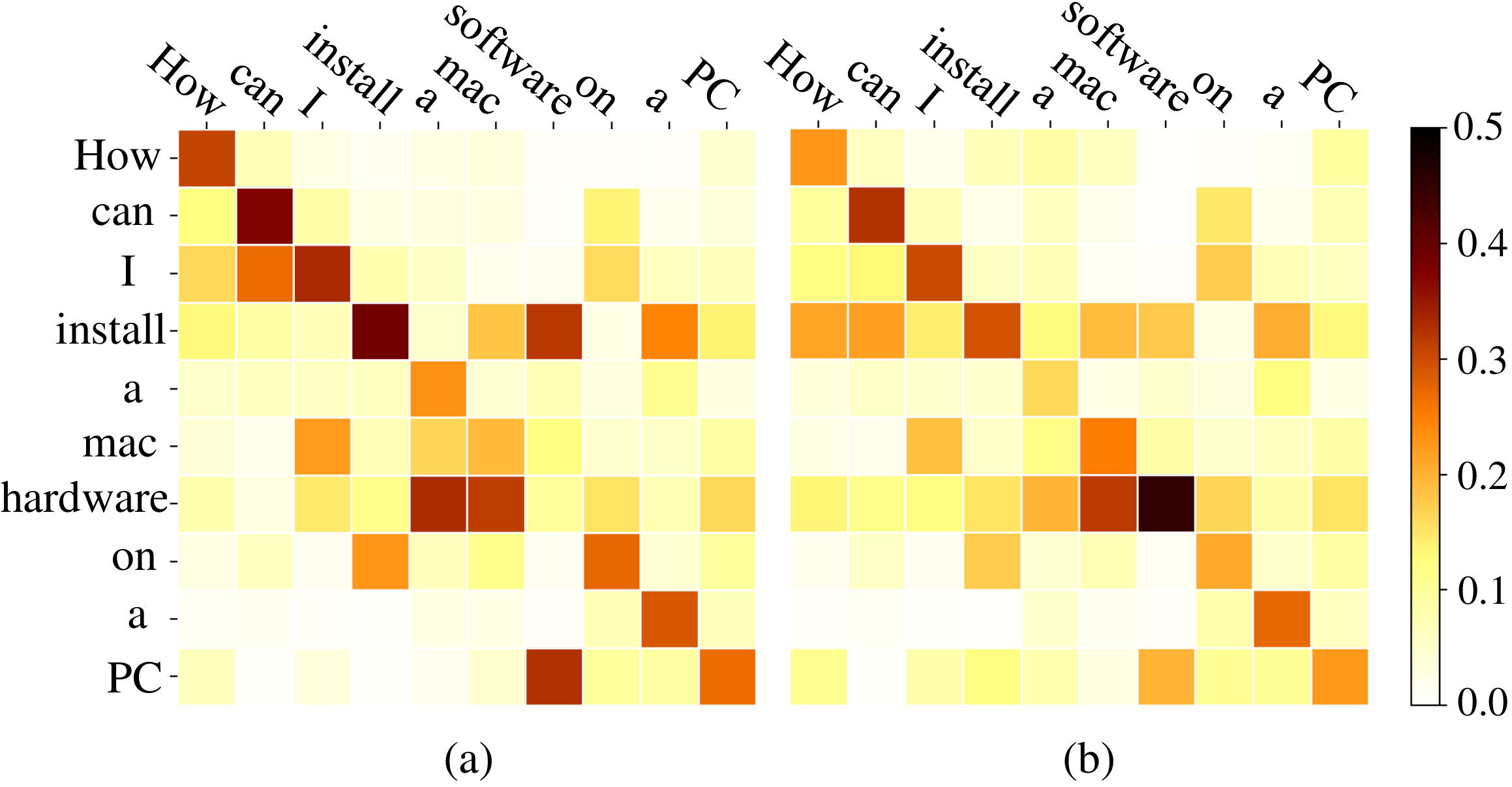}
\caption{\label{fig:heatmap}Distribution of BERT (a) and our method (b).}
% of BERT (a) and our method (b).
\end{figure}

\noindent \textbf{Attention Distribution.} To verify the fusion effect of subtraction-based  attention on the difference information, we display the weights distribution of BERT and DABERT in Figure \ref{fig:heatmap} for comparison.
It can be seen that the attention distribution after dual attention becomes more reasonable, especially the attention weight between "hardware" and "software" increases significantly.
This reveals that DABERT pays more attention to different parts of sentence pairs rather than the same words.

\section{Related Work}

\textbf{Semantic Sentence Matching} is a fundamental task in NLP. Thanks to the appearance of large-scale annotated datasets \cite{bowman2015large,williams2017broad}, neural network models have made great progress in SSM \cite{qiu2015convolutional}, mainly fell into two categories. The first \cite{conneau2017supervised,choi2018learning} focuses on encoding sentences into corresponding vectors without cross-interaction and applies a classifier to obtain similarity. The second \cite{wang2017bilateral,chen2016enhanced,liang2019asynchronous,xue2023dual,liu2023time} utilizes cross-features as an attention module to express the word- or phrase-level alignments of two texts, and aggregates it into prediction layer to acquire similarity. Recently, the pre-training paradigm has achieved great results in SSM. 
Some work attempt to introduce other methods to enhance pre-trained models. For example, SemBERT \cite{song-etal-2022-improving-semantic} explicitly absorbs contextual semantics over a BERT backbone. AMAN \cite{liang2019adaptive} uses answers knowledge to enhance language representation. UER-BERT \cite{xia2021using} injects synonym knowledge to enhance BERT. Syntax-BERT \cite{bai2021syntax} also integrates the syntax tree into transformer models.

\textbf{Robustness} Although neural network models have achieved human-like or even superior results in multiple tasks, they still face the insufficient robustness problem in real application scenarios \cite{gui2021textflint}. Tiny literal changes may cause misjudgments.
Therefore, recent work starts to focus on robustness research from multiple perspectives. TextFlint \cite{gui2021textflint} incorporates multiple transformations to provide comprehensive robustness analysis. \citet{li2021searching} provide an overall benchmark for current work on adversarial attacks. And \citet{liu2021explainaboard} propose a more comprehensive evaluation system and add more detailed output analysis indicators.

\section{Conclusions}
\label{sec:bibtex}

In this paper, we propose a novel Dual Attention Enhanced BERT (DABERT), which can efficiently aggregate the difference information in sentence pairs and soft-integrate it into a pretrained model. Based on BERT's powerful contextual representation capability, DABERT enables the model to learn more fine-grained comparative information and enhances the sensitivity of PLMs to subtle differences. Experimental results on 10 public datasets and robustness dataset show that our method can achieve better performance than several strong baselines. Since DABERT is an end-to-end training component, it is expected to be applied to other large-scale pre-trained models in the future.

\bibliography{custom}
\bibliographystyle{acl_natbib}

% \appendix

% \section{Example Appendix}
% \label{sec:appendix}

\end{document}